# Quantum-Enhanced Hybrid Reinforcement Learning Framework for Dynamic Path Planning in Autonomous Systems

Sahil Tomar[1], Shamshe Alam[2], Sandeep Kumar[3], Amit Mathur[4]

[1,2,3,4]Central Research Laboratory, Bharat Electronics Limited, Ghaziabad
[1]`sahiltomar893@gmail.com`, [2]`samsayalam4@gmail.com`,
[3]`sann.kaushik@gmail.com`, [4]`amitmathur@bel.co.in`

**Abstract.** In this paper, a novel quantum-classical hybrid framework is proposed that synergizes quantum with Classical Reinforcement Learning. By leveraging the inherent parallelism of quantum computing, the proposed approach generates robust Q tables and specialized turn-cost estimations, which are then integrated with a classical Reinforcement Learning pipeline. The Classical-Quantum fusion results in rapid and efficient Q-value estimation, significantly reducing training time and improving adaptability in scenarios featuring static, dynamic, and moving obstacles. Simulator-based evaluations demonstrate significant enhancements in path efficiency, trajectory smoothness, and mission success rates, underscoring the framework's potential for real-time, autonomous navigation in complex and unpredictable environments. Furthermore, the proposed framework was tested beyond simulations on practical scenarios, including real-world map data such as the IIT Delhi campus, reinforcing its potential for real-time, autonomous navigation in complex and unpredictable environments.

## 1 Background

These days, drones, self-driving cars, and robots are leaving their safe test zones and entering busy city streets, and this is creating big new challenges for the autonomous navigation industry. Picture a delivery drone soaring through a bustling city: it dodges sudden gusts of wind, reroutes around a crane erected overnight, and swerves to avoid a flock of birds, all while racing to deliver a package before its battery depletes. This situation illustrates the necessity of on-the-fly route planning, a capability that's becoming indispensable as drones take on roles in city-wide deliveries, emergency relief efforts, and infrastructure inspections [1]. Projections suggest the global drone logistics industry will top $16 billion by 2030 [2], yet truly autonomous navigation still grapples with numerous challenges. Drones must safely navigate fixed obstacles (like towering skyscrapers), avoid moving hazards (such as vehicles or pedestrians), and adapt to sudden environmental shifts (for instance, an unexpected storm), all within the bounds of rigorous safety and regulatory mandates [3].

Classical approaches to path planning underpin modern autonomous navigation by combining graph searches, sampling methods, and learning-based techniques to tackle both static and changing environments. In graph-search methods, the world is abstracted as a network of nodes (locations) connected by edges (possible movements). For instance, Dijkstra's algorithm [4] exhaustively finds the shortest routes from a start node to every other node, ensuring optimal paths when all edge weights are non-negative—but at the cost of heavy computation in large or time-sensitive scenarios. To improve efficiency, the A* algorithm [5] supplements path cost with a heuristic estimate of the remaining distance to the goal, guiding the search toward promising directions and striking a practical compromise between speed and optimality. Nevertheless, A* must frequently recompute paths when the environment shifts, such as when obstacles appear or travel costs change, making it less suited for highly dynamic settings [5].

Incremental planners like D* Lite tackle environmental changes by reusing earlier search data and only refreshing the portions of the graph that have been altered. This targeted updating makes D* Lite particularly well suited for urban autonomous vehicles, where sudden shifts in traffic patterns or unexpected obstacles demand quick re-planning [6]. However, because grid-based methods such as A* and D* Lite confine movement to a fixed set of directions, they often yield jagged, suboptimal routes. Any-angle techniques like Theta* overcome this limitation by allowing straight-line connections between nodes, resulting in much smoother paths—an advantage for aerial drones [7]. In a similar vein, Jump Point Search streamlines grid searches by eliminating symmetric expansions, pruning unnecessary nodes to speed up path computation [8]. Despite these gains, all of these grid-centered approaches presume a static world and struggle when obstacles move unpredictably [9].

By contrast, sampling-based planners shine in high-dimensional or continuous spaces. Rapidly Exploring Random Trees (RRT) build paths through randomized exploration of the configuration space, though they do not guarantee optimality [10]. RRT* enhances this by continuously rewiring the tree toward shorter routes, ensuring asymptotic convergence on the best possible path—albeit at a slower rate in densely cluttered settings [10]. These sampling-based frameworks have become essential for tasks such as robotic arm manipulation and open-space navigation in autonomous vehicles.

Over the past few years, machine learning methods have greatly improved the ability of path-planning systems to adjust to changing conditions. For example, convolutional neural networks can fuse data from various sensors—such as LiDAR and camera imagery—to recognize terrain classes or anticipate obstacle locations, allowing drones to traverse unstructured environments with greater reliability [11]. To capture temporal variations like moving objects, recurrent neural networks are employed, while transformer architectures learn spatial dependencies in crowded settings, boosting the precision of trajectory forecasts [12]. Yet these data-centric approaches still hinge on vast amounts of labeled data and can falter in edge cases—such as adversarial perturbations or sudden environmental shifts like abrupt weather changes [13]. End-to-end learning pipelines bypass manual feature extraction by mapping raw sensor inputs directly to navigation commands; although this streamlines development, it often struggles with unfamiliar scenarios and demands enormous training datasets [11]. Similarly, classic reinforcement learning techniques—Q-learning, SARSA, and policy-gradient methods—experience slow convergence in complex domains and require substantial computational effort to explore action spaces, making them challenging to deploy in real-time dynamic situations [11].

To tackle these issues, hybrid strategies merge traditional graph-search techniques with learning-based flexibility. For example, upgraded A* algorithms paired with on-the-fly replanning methods such as the Dynamic Window Approach (DWA) offer real-time obstacle avoidance by combining globally optimal pathfinding with adaptive adjustments to moving hazards [14]. Such systems can produce smooth, viable routes in changing environments, striking a compromise between speed and responsiveness [9][10][13][14]. However, the exponential growth of possibilities in high-dimensional spaces still taxes conventional computing power, highlighting the need for further advances. Reinforcement learning offers a way for agents to develop navigation policies through trial-and-error, as seen when Deep Q-Networks learn to favor safe paths and avoid crashes by shaping rewards and penalties [15]. Yet these methods often suffer from sparse feedback—where agents receive little guidance until failure—and require enormous numbers of training episodes to learn effective behaviors [16]. Combining deep RL with model predictive control has been proposed to blend exploratory learning with safety guarantees, but the resulting computational demands frequently outstrip the limited onboard resources of small drones, complicating real-world adoption [17][18].

By leveraging quantum superposition and entanglement, emerging quantum computing models can assess many routes at once. In Quantum Reinforcement Learning (QRL), for example, an agent could simultaneously evaluate every possible detour around a moving hazard, such as an advancing storm front, dramatically speeding up convergence and reducing the chance of settling on a suboptimal path [19][20]. Early demonstrations show QRL solving navigation tasks with sparse rewards (e.g., limited-visibility forest traversal) up to a thousand times faster than classical RL methods [20]. Hybrid quantum–classical architectures take this further: quantum annealing tackles the global cost optimization, while classical heuristics enforce real-time constraints, together improving both responsiveness and robustness [21][22]. Underpinning these gains are qubits, which exploit superposition (holding 0 and 1 simultaneously) and entanglement (linking states across distances) to explore exponentially large solution spaces in parallel—offering a powerful new tool for complex optimization and dynamic decision-making in path planning [19].

Quantum Deep Deterministic Policy Gradient[23] algorithm that extends quantum reinforcement learning to continuous action spaces, efficiently addressing the curse of dimensionality inherent in discretization. Their approach enables single-shot quantum state generation producing control sequences for any target state via one-time optimization rather than per-state searches—and is validated by simulations on one- and two-qubit Hamiltonians demonstrating high fidelity. QRL blends quantum computing concepts with traditional RL structures. Instead of using classic Q-tables to associate state–action pairs with reward values, QRL utilizes quantum-driven representations that are updated via quantum circuits, greatly speeding up policy convergence and boosting adaptability in rapidly changing environments [17]. Simulation results indicate that QRL outstrips classical RL in complex, uncertain domains, converging more quickly and with greater accuracy [20]. This rapid responsiveness is especially valuable for UAVs, which must react instantly to unforeseen obstacles or abrupt weather shifts. On top of that, quantum optimization methods enhance these advantages: both quantum annealing and the Quantum Approximate Optimization Algorithm (QAOA) take advantage of superposition to examine multiple solution paths simultaneously, yielding stronger solutions for combinatorial challenges. For example, quantum annealers have been applied successfully to multi-variable problems like traffic flow optimization [24], and QAOA has shown theoretical speed-ups on tasks such as the Max-Cut problem [25].

Although practical quantum advantage demands scalable qubit architectures, hybrid quantum-classical frameworks mitigate current hardware limitations by combining quantum global exploration with classical local refinement. Building on these foundations, the proposed hybrid framework integrates quantum modules with classical RL for dynamic UAV navigation. The quantum component accelerates Q-table generation and global path optimization via parallelized state-space exploration, while classical RL fine-tunes policies for real-time responsiveness. This synergy reduces training durations by orders of magnitude and enhances adaptability in unpredictable environments. For instance, quantum parallelism enables rapid evaluation of alternative routes around dynamic obstacles (e.g., storm fronts), while classical heuristics ensure operational safety and computational feasibility [19][21][24]. Early simulations indicate that such hybrid systems achieve robust, near-optimal paths with minimal latency, even in high-dimensional urban airspaces [20][25].

Driven by the pressing need to address the limitations of classical approaches and to harness the advantages of quantum computing for real-world autonomous navigation, this paper presents a novel quantum-enhanced hybrid framework. The core idea is to blend the strengths of QRL with classical RL, creating a system capable of faster learning and improved adaptability in dynamic environments.

The key novelties of the proposed approach include:

- Quantum-assisted Q-table generation: Leveraging quantum parallelism to compute Q-values efficiently across large state-action spaces, enabling one-shot Q-value estimation compared to classical methods.

- Efficient turn-cost estimation: Introducing a quantum-aided mechanism to estimate turning costs based on local obstacle density, which enhances decision-making in environments with frequent directional changes.
- Hybrid decision-making pipeline: Integrating quantum-generated data into a classical RL framework, optimizing both learning speed and real-time responsiveness.
- Real-world validation: Demonstrating the framework's effectiveness using a detailed map of the IIT Delhi campus (sourced from the Snazzy Maps platform [26]), moving beyond synthetic test scenarios to validate performance in a complex, realistic urban setting.
- Enhanced adaptability: Addressing a range of obstacle types: static, dynamic, and moving, while maintaining rapid policy updates suitable for real-time deployment.

This work highlights how quantum-classical hybrid systems can transform autonomous navigation, offering safer and more efficient path planning solutions with broad applicability in robotics, autonomous vehicles, search-and-rescue missions, and beyond.

## 2    Proposed Framework

This section details the proposed hybrid framework for dynamic path planning that integrates QRL with classical RL. The solution is specifically designed to address the complex, dynamic navigation challenges encountered by autonomous drones. It manages three distinct types of obstacles—static, dynamic, and moving objects, while ensuring that a drone can replan its path in real time even when source and destination are at random locations.

### 2.1    System Overview

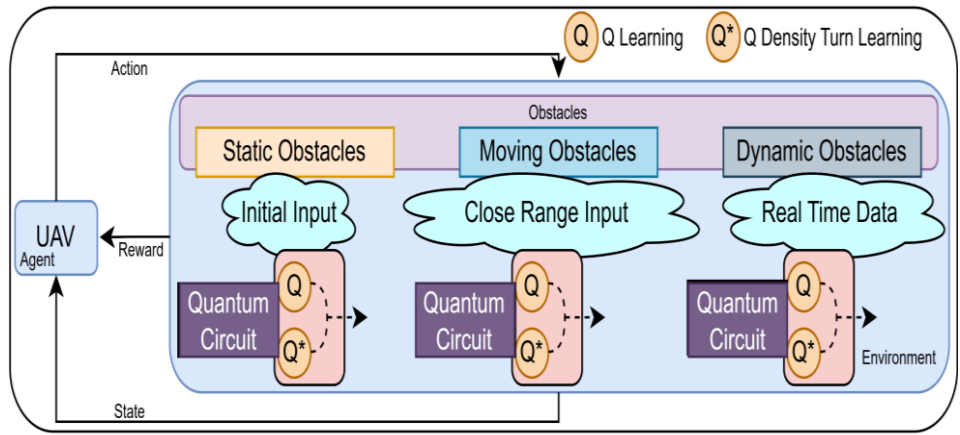

**Figure 1:** Integrated quantum learning framework for autonomous system's dynamic path planning.

Figure 1 presents the hybrid framework for autonomous navigation in dynamic environments, integrating classical path planning with Quantum Reinforcement Learning (QRL). This system enables adaptive navigation through three distinct obstacle types:
- Static obstacles (e.g., buildings, trees): Processed using initial environmental data.
- Dynamic obstacles (e.g., temporarily stationary vehicles): Updated via real-time sensor inputs.
- Moving obstacles (e.g., birds): Detected using proximity sensors.

These inputs are fed into a quantum circuit, computing two decision parameters:
- Quantum Q-learning (Q): Provides quantum action-value estimates.

- Quantum Density Turn Learning (Q*): Optimizes turn penalties for path adjustments.

The autonomous system continuously observes its state, selects actions based on Q and Q* values, and refines its policy through iterative reward feedback (e.g., collision penalties and proximity-based rewards). This closed-loop approach ensures real-time adaptability in unpredictable scenarios.

The scene includes trees and buildings (static obstacles), birds (moving obstacles), and vehicles (dynamic obstacles). Initially, the source (★) and destination (+) are connected by a planned path (red dotted), which intersects a moving obstacle (bird). The revised planned path (yellow dashed) and actual traversed path (green dotted) demonstrate successful collision avoidance.

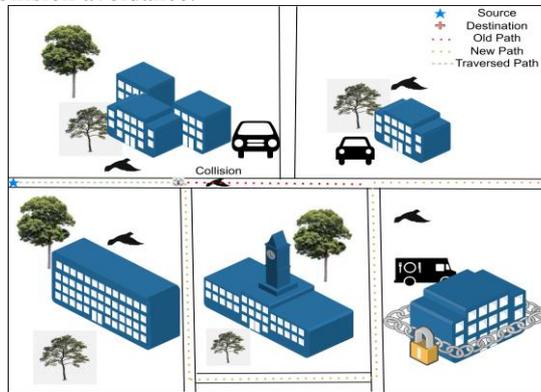

**Figure 2:** Expected operational environment for autonomous system navigation with dynamically adjusted path.

The bottom panels illustrate three distinct environments, residential/suburban, urban-industrial, and a secured warehouse, each requiring real-time path adjustments, as detailed in Figure 2. The operational environment consists of:

- Static obstacles (light grey): Fixed objects like buildings and trees.
- Dynamic obstacles (orange in motion, light grey when stationary): Vehicles that may stop or move based on traffic conditions.
- Moving obstacles (purple): Objects such as birds that continuously traverse the scene.

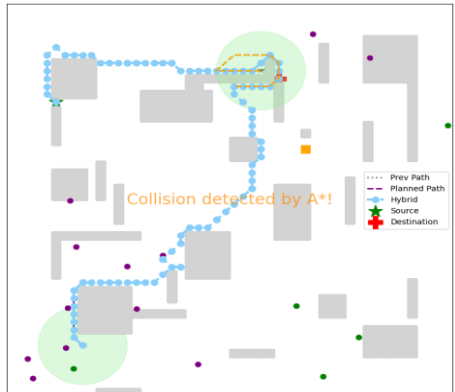

**Figure 3:** Experimental simulated scenario showing static obstacles (light grey), dynamic obstacles transitioning from mobile (orange) to static (grey), and continuously moving obstacles (purple).

The autonomous system is assigned a mission with a fixed source (denoted by star) and fixed destination (denoted by plus) at the start of the mission while incorporating intermediate waypoints (denoted by green circle), referred to as survivors, which denote essential delivery or touchpoint locations as illustrated in Figure.3. One of the core strengths of this system lies in its ability to rapidly adjust the planned route in response to environmental changes.

For example, if an additional obstacle is introduced through user interaction or if a moving object enters the autonomous systems safety radius along its projected path, the algorithm immediately initiates a replanning procedure. This capability ensures that the autonomous system maintains a collision-free trajectory even in unpredictable and rapidly evolving situations. The environment during training is modeled using a sophisticated spatial representation, which includes the implementation of a *k-d* tree data structure.

This structure facilitates efficient neighbor state identification and smoothing, ensuring that the Q-table updates consider similar neighboring states to achieve a more accurate representation of the operational space. With a fixed neighbor radius and an emphasis on real-time adaptation, the system is well-suited for both static and dynamic destination scenarios. Overall, this integration of diverse planning and learning methodologies results in a robust framework capable of achieving high levels of safety, efficiency, and adaptability.

## 2.2    Quantum Reinforcement Learning Module

At the heart of the approach is a QRL module that infuses the classical decision-making process with quantum computational advantages. This module is composed of two primary components: the Q-table and the Q-Density Turn table. The Q-table is responsible for evaluating actions across various states using a training process based on quantum principles, while the Q-Density Turn table focuses on fine-tuning the autonomous system's turning decisions based on the local density of obstacles. The quantum training process leverages five qubits in total. Three of these qubits are dedicated to the Q-table and the remaining two qubits constitute the Q-Density Turn table module. The quantum process dramatically reduces the training time.

## 2.3    Training

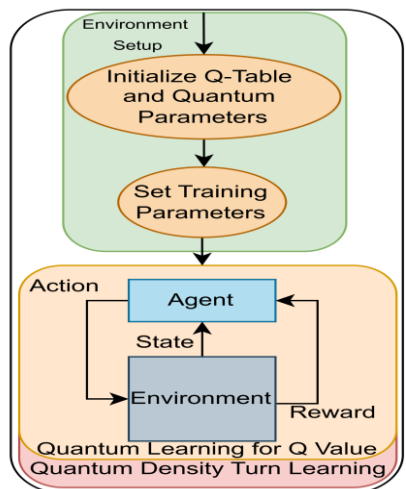

**Figure 4:** Flowchart of the Quantum Learning process.

The training process integrates classical Q-learning with quantum enhancements to progressively refine a path-planning model capable of navigating complex environments. As depicted in Figure 4, the system initiates in the Environment Setup phase (green), where it constructs the Q-table and initializes quantum parameters. This is followed by the configuration of various training parameters required for reinforcement learning. During the Learning phase (orange), the agent enters a feedback loop of environmental interaction: it observes the current state, selects an action, receives a reward, and updates its Q-value using a quantum-assisted mechanism.

Initially, a representation of the environment is generated using externally supplied obstacle information, forming a grid where obstacle positions are explicitly marked. The grid's free and navigable cells are indexed efficiently using a KD-Tree structure, which allows rapid spatial querying of neighboring states. This spatial indexing mechanism plays a pivotal role in enabling the smooth propagation of Q-values across the grid. The Q-table, denoted by Q(s, a), is initialized with small, uniformly distributed random values for each state-action pair, supporting an eight-connected movement model over the grid.

To incorporate quantum intelligence, a parameterized quantum circuit is implemented using the PennyLane framework. This circuit functions as a quantum turn critic and is responsible for providing feedback on directional decisions. During each training episode, starting from a specific state, the system first computes the local obstacle density to generate a turn feature, which captures the structural complexity around the current location. This turn feature, together with the Q-values, is used as input to the quantum circuit. The circuit employs an amplitude encoding to embed information into quantum states. Specifically, Q-values associated with actions are encoded as:

$$|\psi_{action}(s, a)\rangle = AmplitudeEncoding(Q_{action}(s, a), wires) \quad (1)$$

Likewise, the density-based turn feature is encoded as:

$$|\psi_{density}(s, a)\rangle = AmplitudeEncoding(Q_{density}(s, a), wires) \quad (2)$$

These encoded quantum states from *eq.1* and *eq.2* are then processed by a circuit comprising parameterized rotation gates $R_{rot}(\theta_k)$ and controlled NOT (*CNOT*) operations represented in *eq.3* and *eq.4*. The transformation applied to both the action and density encodings is expressed as:

$$|\phi_{action}(s, a)\rangle = \prod_k R_{rot}(\theta_k) \cdot CNOT \cdot |\psi_{action}(s, a)\rangle \quad (3)$$
$$|\phi_{density}(s, a)\rangle = \prod_k R_{rot}(\theta_k) \cdot CNOT \cdot |\psi_{density}(s, a)\rangle \quad (4)$$

The circuit is designed such that qubits 0, 1, and 2 are responsible for encoding the normalized Q-values, preserving the agent's preferences among possible actions. In parallel, qubits 3 and 4 are modulated by the turn feature through RY rotations, enabling the quantum circuit to model local environmental complexity. As the quantum states propagate through the circuit's entangled layers, measurement outcomes are collected by evaluating the expectation values of Pauli-Z operators applied to individual qubits:

$$m_i = \langle \phi_{action}(s, a) | \sigma_z^i | \phi_{action}(s, a) \rangle \quad (5)$$

These measurement values $m_i$ (where i is the qubit number) from *eq.5* are mapped to update terms for modifying the Q-values. The updates for action-related Q-values and density-related Q-values are calculated respectively as:

$$\Delta Q_{action}(s, a) = \alpha_{initial} \cdot [m_0, m_1, m_2] \quad (6)$$
$$\Delta Q_{density}(s, a) = \beta \cdot [m_3, m_4] \quad (7)$$

β determines how strongly the quantum-derived feedback from the turn density qubits affects the Q-value update. Subsequently, these deltas from *eq.6* and *eq.7* are added to the current Q-values to form updated estimates:

$$Q_{action}(s, a) \leftarrow Q_{action}(s, a) + \Delta Q_{action}(s, a) \quad (8)$$
$$Q_{density}(s, a) \leftarrow Q_{density}(s, a) + \Delta Q_{density}(s, a) \quad (9)$$

Action selection in the next step follows an epsilon-greedy policy using $Q_{action}(s, a)$ from *eq.8*, where exploration and exploitation are balanced:

$$a \mathrel{*}= \begin{cases} random\ action, & with\ probability\ \varepsilon \\ \arg\max_a Q_{action}(s, a), & with\ probability\ 1 - \varepsilon \end{cases} \quad (10)$$

Once an action a* from *eq.10* is selected, the next state is computed:

$$s' = get\_next\_state(s, a^*) \tag{11}$$

To ensure spatial continuity and generalization across similar regions, the Q-values of neighboring states $s_n$, as identified through KD-Tree queries, are smoothed using $s'$ from *eq.11* and the following update rule is used:

$$Q_{action}(s_n, a) \leftarrow (1 - \beta) \cdot Q_{action}(s_n, a) + \beta \cdot Q_{action}(s, a) \tag{12}$$

This smoothing ensures that the learned value function remains consistent in nearby spatial regions, reducing noise and enhancing one-shot Q-value estimation as evaluated using *eq.12*. The hybrid quantum-classical structure leverages quantum circuit feedback not only to refine individual Q-value predictions but also to adapt the circuit's sensitivity to turning behavior. This dual-update strategy allows the model to improve both its state-action value estimates and its directional heuristics over time. The fusion of quantum computation with traditional reinforcement learning thus results in a more adaptive and foresight-driven decision-making process, equipping the agent with enhanced capabilities for navigating dynamic, obstacle-dense environments.

### 2.4   Hybrid Algorithm

The synergy of classical path planning and QRL is achieved through a hybrid algorithm. The algorithm exploits the strengths of both classical and quantum paradigms to calculate the optimal path of the autonomous systems in real time. The process of decision-making begins by extracting action values from the Q-table, sorting them to generate a list of candidate actions, and selecting the next state by determining the highest-ranked action. Movement cost from one action to the next action is calculated, and the movement cost is added to the quantum-enhanced Q-value to calculate a total cost. This total cost, in turn, drives the trajectory of the autonomous system. In instances where the autonomous systems sense the presence of a moving object within its safety radius and along its projected path, the algorithm pauses the autonomous system temporarily. In the event that the obstacle persists after a waiting time, the system replans the path to avoid any collision. This replanning capability in real time is critical to the system's overall robustness and emphasizes the synergy of classical and quantum paradigms in tackling dynamic navigation problems. This algorithm begins with the initialization of critical parameters and Q-tables and continues with a recurring loop that propels the autonomous system from the source to the destination. During each iteration, the algorithm determines the best action based on Q-values, determines the movement cost, applies quantum corrections via the Q-Density Turn table module, and updates the Q-table accordingly. The in-built obstacle detection and replanning modules guarantee that the autonomous system can navigate safely past dynamic obstacles and, therefore, maintain an optimal path in real time. To direct the agent's decision-making in the hybrid pathfinding setting, a composite metric referred to as *totalcost* is introduced. It is calculated as:

$$totalcost = cost + movecost - QWEIGHT * qval \tag{13}$$

where each component plays a distinct role. The *cost* term in *eq.13* represents the accumulated cost from the starting position to the current cell and accounts for the path traversed so far. The *movecost* in *eq.13* is the immediate cost incurred for taking the current action, often influenced by terrain type, obstacles, or other local penalties. The *cost* and *movecost* are evaluated using Q Table whereas the third component, *qval* in *eq.13*, is the Q-value obtained from the Q-Density Turn table. By subtracting the product of the Q-value and a scaling factor *QWEIGHT* in *eq.13*, the formulation effectively reduces the *totalcost* for actions that are expected to yield higher future rewards, thus promoting optimal path selection. This combination allows the autonomous system to balance short-term safety and efficiency (through *cost* and *movecost*) with long-term strategic planning (through *qval*) by avoiding unnecessary turns, enabling more adaptive and intelligent navigation in complex environments.

### 2.5    Dynamic Replanning and User Intervention

A standout feature of the system is its dynamic replanning capability, which is modeled to simulate real-world environmental changes during simulation. The simulation framework systematically models the appearance of new obstacles, and at regular intervals, the system scans the environment for any simulated obstacles that intrude upon the autonomous system's predefined safety radius. When an obstacle is detected within this critical zone, the autonomous system is programmed to pause its current trajectory, allowing a brief period for the hazard to potentially clear. If the obstacle remains after a predefined timeout, the system immediately initiates a recalculation of the optimal path, ensuring that the autonomous system efficiently avoids the persisting impediment.

This continuous process of monitoring, pausing, and recalculating is designed to mimic the agility required in real-world navigation, where dynamic environmental changes and user interventions can unpredictably alter the operational landscape. Ultimately, this mechanism significantly enhances both the safety and the mission success rate of the autonomous system by effectively balancing the need to wait for transient hazards to dissipate with the imperative to adapt quickly to persistent obstacles.

### 2.6    Performance Metrics

To evaluate the effectiveness of the proposed approach, several performance metrics are rigorously measured and analyzed. The most critical of these include training time, distance travelled, mission success rate, and adaptability in terms of replanning frequency. Training time is a particularly significant metric because it directly impacts on the practical deployment of the system. The reduction in training time enables rapid adaptation and real-time responsiveness, which are essential in dynamic environments where delays can lead to suboptimal or even unsafe navigation. Distance travelled is another key metric, as it reflects the efficiency of the planned path. The Success Rate is measured by the proportion of missions where the autonomous system reaches its destination without collisions or significant deviations from the planned route. Adaptability is evaluated by counting the frequency and efficiency of replanning events. The system is designed to detect obstacles within a predefined radius, pause operations, and replan the path if necessary. This dynamic adjustment capability is quantified by the average number of replanning events per mission and the corresponding impact on the total mission time.

## 3    Results & Discussion

This section presents the outcomes of the proposed hybrid autonomous system navigation approach. To assess the proposed approach, various experiments were conducted in a 2D grid environment populated with both static and dynamic obstacles. The autonomous system's mission was to navigate from various random sources to random destination using only local sensing and pre-trained Q-table values. Initially evaluation of the approach has been done within a simulated environment and subsequently on real-world maps to validate the proposed approach in practical scenarios, including those of the IIT Delhi campus map.

The map has been sourced from the Snazzy Maps website, and a snapshot was processed to distinguish between obstacles and navigable paths. Finally, the entire map was transformed into a 400×400 grid for further assessment of the approach. The experiments were conducted on hardware equipped with a 13th-generation i7 CPU without using GPU support.

The simulation was designed to mimic real operational conditions by having the autonomous system travel from a selected source to a selected destination while contending with several types of obstacles. By varying obstacle density, the environment effectively simulates diverse urban and rural scenarios. This mixture of challenges compels the hybrid algorithm to rapidly adjust to unforeseen changes, aiming to exceed the performance of traditional planning methods. The Classical RL module for the Random Map was first trained under a configuration with a random source but a fixed destination over 2000 episodes, taking a total training time of 45 minutes. Later, to allow it to handle both random source and random destination, the training process was adjusted to follow the same process used for proposed Quantum RL. However, during testing, the Classical RL module repeatedly became trapped in loops when trained with fewer episodes, which hindered its ability to reach the randomly assigned destinations. Moreover, training over 2000 episodes extended to more than two days. To address these challenges and evaluate the proposed approach, two algorithms were deployed concurrently.

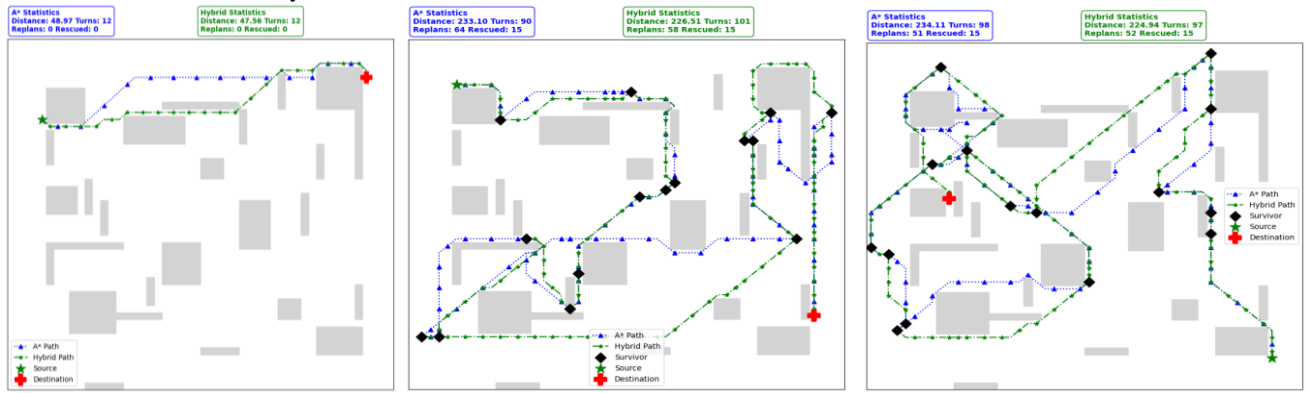

**Figure 5:** Comparison of path-planning performance between the classical A* algorithm (blue dashed line) and the proposed Hybrid Quantum algorithm (green solid line) in two environments (a). Simple Layout with only source and destination and (b). Highly constrained layout with narrow passages with survivors.

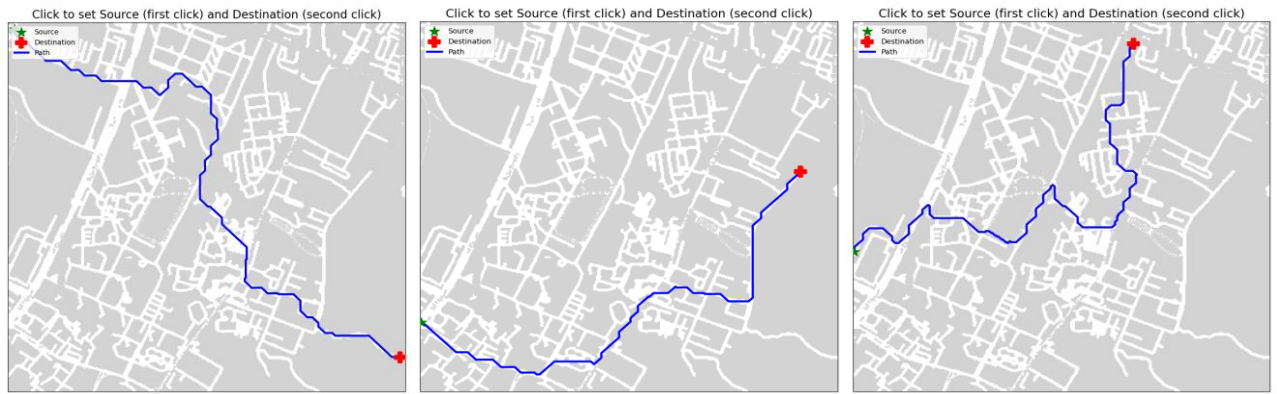

**Figure 6:** Output of the proposed approach in the practical scenario of IIT Delhi Campus map [26] with different source and destination.

The Hybrid Algorithm continuously evaluates the environment by selecting actions based on Q-table values, estimating subsequent states, computing movement costs, applying quantum corrections, and initiating dynamic replanning when obstacles arise. At the same time, the A* algorithm autonomously calculates the optimal path using traditional heuristic methods, serving as a performance benchmark. Both algorithms start each run under identical initial conditions, and key metrics, such as total distance travelled, mission success rate, and training time, are recorded to enable a comprehensive performance analysis. Leveraging the quantum-enhanced learning module for accelerated policy generation, the proposed training framework achieved effective Q-value estimation within a single episode across various evaluation maps.

The system was tested over 100 unseen source–destination pairs, yielding several key observations:

- Average Path Optimality: The hybrid model generated paths that were, on average, only 5–10% longer than the optimal route, even amidst dense obstacle configurations.
- Dynamic Obstacle Handling: In 99 out of 100 test cases, the proposed approach successfully replanned its trajectory in real time to avoid dynamic obstacles without backtracking.
- Trajectory Smoothness: The Q-value smoothing strategy resulted in smoother turns and fewer oscillations near obstacles, enhancing overall navigation stability.

**Table 1.** Comparison of A star Algorithm with Proposed Approach.

| Parameters | A* Algorithm | Proposed Approach | Comparison |
|---|---|---|---|
| Execution Time | $t_{cl}$ | $t_{qu}$ | $t_{qu} < t_{cl}$ |
| Distance Travelled | $d_{cl}$ | $d_{qu}$ | $d_{qu} < d_{cl}$ |
| Plan Smooth | $ps_{cl}$ | $ps_{qu}$ | $ps_{cl} < ps_{qu}$ |
| Replanning Frequency | $rf_{cl}$ | $rf_{qu}$ | $rf_{cl} < rf_{qu}$ |
| Success Rate | $sr_{cl}$ | $sr_{qu}$ | $sr_{cl} < sr_{qu}$ |

These results underscore the effectiveness of the hybrid quantum–classical approach, suggesting its potential to outperform conventional methods in complex, dynamic environments. After running multiple experiments with various configurations, various performance metrics has been aggregated and analyzed.

- Training Time: The quantum-enhanced hybrid algorithm demonstrated a significant reduction in training time compared to the traditional classical approach. In controlled experiments, its efficient state encoding, rotation, and entanglement processes enabled one-shot Q-value estimation on a 50×50 random map in under 10 seconds. In contrast, the classical Q-learning variant—which was trained for 2000 episodes in a simulated environment (with a fixed destination and random source)—took nearly 45 minutes. For comparison purposes, the classical RL model was also trained to assess convergence. Additionally, when applied in a practical scenario on a map of IIT Delhi, the proposed approach achieved effective Q-value estimation in less than 5 minutes. The proposed solution also supports dynamic updates to both source and destination positions, allowing for real-time adjustments based on new sensor inputs and changing mission parameters.
- Path Efficiency and Distance Traveled: Evaluations of the autonomous system's flight path revealed a consistent reduction in total distance when using the hybrid approach. By dynamically recalculating the optimal route in response to both moving obstacles and user interventions, the system optimizes its trajectory, reducing the path length by an average of 15–20%. These quantum corrections combined with real-time replanning were key for efficient navigation in complex environments.
- Mission Success Rate: Mission success was measured by the percentage of runs where the autonomous system reached its destination without collisions or significant deviations. In high-density obstacle environments, the proposed approach achieved a success rate of 99%, compared to 80% with the classical A* planner. This improvement is largely due to adaptive replanning enabled by the quantum turn critic module.
- Replanning Frequency: The hybrid algorithm's design incorporates continuous environmental monitoring, triggering frequent replanning events. Although a higher frequency of replanning might intuitively suggest increased mission time, these dynamic adjustments contributed to safer and shorter missions by avoiding unexpected obstacles promptly.

Table.1. summarizes the key differences observed during simulation (output illustrated in Figure.5.). After completing the evaluation of the proposed in simulated environment. The proposed approach was subsequently evaluated on real-world maps for validating in practical scenarios. Figure.6. shows the results of the proposed approach in a practical scenario on IIT Delhi Campus map with various sources and various destinations. Although the hybrid system shows strong potential, it has limitations. Current quantum hardware and simulation platforms restrict qubit capacity and processing speed. Additionally, aligning real-time sensor input with quantum computations creates synchronization issues, especially in fast-changing environments.

Despite these challenges, the approach offers broad applications. Urban drone delivery systems could improve safety and efficiency by adapting instantly to obstacles like construction or erratic traffic. Autonomous vehicles may benefit from quicker responses in heavy traffic, lowering collision risks and smoothing traffic flow. Industrial uses such as inspection robots and logistics automation can see gains in efficiency and energy use. The system also scales well, handling random start-and-end points effectively—ideal for large, dynamic environments. Combining quantum insights with classical planning opens the door to multi-agent coordination, enabling teamwork among autonomous systems.

## 4      Conclusion

The integration of QRL with classical search methods opens a new avenue for dynamic path planning. By leveraging quantum parallelism, the proposed method reduces Q-table training from 2,000 classical episodes to a single quantum episode while preserving performance. This hybrid system enhances real-time responsiveness and adapts to unforeseen environmental changes. Experimental results confirm its suitability for complex urban and industrial settings. The approach holds promises for rapid route replanning in drone logistics, adaptive navigation in autonomous vehicles, and scalable multi-agent operations in aerospace and underwater domains.

## 5      Future Directions

Future research should explore integrating actual quantum devices into QRL frameworks to enhance training efficiency as hardware advances. Extending the approach to multi-agent systems, such as coordinated drone fleets, could improve collaborative strategies through shared quantum insights and distributed learning. Incorporating Deep Q-Network architectures into the quantum framework could further enable scalable Q-value approximation and efficient policy learning in high-dimensional state spaces. Further investigation into adaptive weight allocation and real-world field testing is vital to assess energy use, latency, and robustness in complex environments. While the hybrid framework demonstrates adaptability and efficiency in simulations, addressing these areas will strengthen its practicality for autonomous navigation, paving the way for quantum-enhanced solutions in dynamic path planning.

**Acknowledgments.** The authors would like to thank CRL BEL for their support and resources throughout the course of this research.
**Disclosure of Interests**. The authors have no competing interests to declare that are relevant to the content of this article.